\documentclass{article}

\usepackage{times}
\usepackage{graphicx} 
\usepackage{subfigure} 
\usepackage{bm}

\usepackage{natbib}

\usepackage{algorithm}
\usepackage{algorithmic}
\usepackage{pdfpages}

\usepackage[accepted,nohyperref]{icml2015}

\usepackage{url}
\usepackage{amsmath}
\usepackage{amssymb}

\newcommand{\given}{\,|\,}
\newcommand{\DistGam}{\text{Gam}}

\icmltitlerunning{Probabilistic Backpropagation}

\begin{document} 

\twocolumn[
\icmltitle{Probabilistic Backpropagation for Scalable \\Learning of Bayesian Neural Networks}

\icmlauthor{Jos\'{e} Miguel Hern\'{a}ndez-Lobato}{jmh@seas.harvard.edu}
\icmlauthor{Ryan P. Adams}{rpa@seas.harvard.edu}
\icmladdress{School of Engineering and Applied Sciences, Harvard University,
            Cambridge, MA 02138 USA}


\icmlkeywords{Bayesian Machine Learning, Neural Networks, Expectation Propagation, Active Learning}

\vskip 0.3in
]

\begin{abstract} 
Large multilayer neural networks trained with backpropagation have recently achieved state-of-the-art results in a wide range of problems. However, using backprop for neural net learning still has some disadvantages, e.g., having to tune a large number of hyperparameters to the data, lack of calibrated probabilistic predictions, and a tendency to overfit the training data.  In principle, the Bayesian approach to learning neural networks does not have these problems. However, existing Bayesian techniques lack scalability to large dataset and network sizes. In this work we present a novel scalable method for learning Bayesian neural networks, called probabilistic backpropagation (PBP).  Similar to classical backpropagation, PBP works by computing a forward propagation of probabilities through the network and then doing a backward computation of gradients.  A series of experiments on ten real-world datasets show that PBP is significantly faster than other techniques, while offering competitive predictive abilities.  Our experiments also show that PBP provides accurate estimates of the posterior variance on the network weights.
\end{abstract} 

\allowdisplaybreaks

\section{Introduction}\label{sec:introduction}

Neural networks (NNs) have seen a recent resurgence of interest due to
empirical achievements on a wide range of supervised learning problems.  In
their typical usage, neural networks are highly expressive models that can
learn complex function approximations from input/output examples
\citep{hornik1989multilayer}.  Part of the success of NNs is due to the ability
to train them on massive data sets with stochastic optimization
\citep{Bottou2010} and the backpropagation (BP) algorithm
\citep{RumelhartHintonWIlliams1986}.  This, along with faster
machines, larger datasets, and innovations such as dropout \cite{srivastava2014dropout} and
rectified linear units \citep{nair2010rectified}, have resulted in successes
for NNs on tasks such as speech recognition \citep{hinton2012deep,Hannun2014},
computer vision \citep{krizhevsky2012imagenet,Wu2015} and natural language
processing \citep{collobert2008unified, Sutskever2014}.

Despite all these successes, there are still some challenges in learning NNs
with backpropagation (BP).  First, there are many hyperparameters in
BP-based stochastic optimization that require specific tuning,
e.g., learning rate, momentum, weight decay, etc., each of which may be
layer-specific.  With large data sets, finding the optimal 
values can take a large amount of time, even with an efficient procedure such
as Bayesian optimization \citep{snoek2012practical}.  Second, in NNs trained
with BP, we can only obtain point estimates of the weights in the network.  As
a result, these networks make predictions that do not account for uncertainty
in the parameters.  However, in many cases these weights may be poorly
specified and it is desirable to produce uncertainty estimates along with
predictions.  Finally, it is common practice to use a  very large NN to
flexibly fit data, and then reign in overfitting using regularization terms,
even if a smaller network would be cheaper and easier to train.

A Bayesian approach to neural networks can potentially avoid some of the
pitfalls of stochastic optimization \citep{mackay1992practical}.  Bayesian
techniques, in principle, can automatically infer hyperparameter values by
marginalizing them out of the posterior distribution or by determining them via
type II maximum likelihood (empirical Bayes). Furthermore, Bayesian methods
naturally account for uncertainty in parameter estimates and can propagate this
uncertainty into predictions.  Finally, Bayesian approaches are often more
robust to overfitting, since they average over parameter values instead of choosing a single point estimate.

Several approaches have been proposed for Bayesian learning of neural networks,
based on, e.g., the Laplace approximation \citep{mackay1992practical},
Hamiltonian Monte Carlo \citep{neal1995bayesian}, expectation propagation
\citep{jylanki2014expectation}, and variational inference
\citep{Hinton93keepingneural}.  However, these approaches have not seen
widespread adoption due to their lack of scalability in both network
architecture and data size.  A notable exception is the scalable variational
inference approach of \citet{Graves2011}.  However, this method seems to
perform poorly in practice due to noise from Monte Carlo approximations within
the stochastic gradient computations.  A different scalable solution based on
expectation propagation was proposed by \citet{Soudry2014}.  While this
method works for networks with binary weights, its extension to continuous
weights is unsatisfying as it does not produce estimates of posterior variance.

We describe a new approach for learning Bayesian neural networks
called \emph{probabilistic backpropagation} (PBP)
that is fast and does not have the disadvantages of previous
approaches. PBP works by propagating probabilities forward through the network to obtain the
marginal likelihood, before propagating backward the gradients of the marginal
likelihood with respect to the parameters of the posterior approximation.  Our
experiments show that PBP is fast, makes accurate
predictions and also produces calibrated estimates of the posterior uncertainty
in network weights.

\section{Probabilistic neural network models}\label{sec:model}

We describe a probabilistic model for data based on a
feedforward neural network. Given data~${\mathcal{D} = \{ \mathbf{x}_n, y_n
\}_{n=1}^N}$, made up of $D$-dimensional feature vectors~${\mathbf{x}_n \in
\mathbb{R}^D}$ and corresponding scalar target variables~${y_n \in
\mathbb{R}}$, we assume that each~$y_n$ is obtained as~${y_n =
f(\mathbf{x}_n;\mathcal{W}) + \epsilon_n}$, where~$f(\cdot ;\mathcal{W})$ is
the output of a multi-layer neural network with connections between consecutive
layers and weights given by~$\mathcal{W}$. The evaluations of this NN are
corrupted by additive noise variables~$\epsilon_n$, where~${\epsilon_n \sim
\mathcal{N}(0,\gamma^{-1})}$.  

The NN has~$L$ layers, where~$V_l$ is the number of hidden units in layer~$l$,
and~${\mathcal{W} = \{ \mathbf{W}_l \}_{l=1}^L}$ is the collection of~${V_l
\times (V_{l-1}+1)}$ weight matrices between the fully-connected layers.  The
$+1$ is introduced here to account for the additional per-layer biases. We
denote the outputs of the layers by vectors~$\{ \mathbf{z}_l \}_{l=0}^{L}$,
where~$\mathbf{z}_0$ is the input layer,~${\{\mathbf{z}_l\}_{l=1}^{L-1}}$ are the
hidden units and~$\mathbf{z}_L$ denotes the output layer, which is
one-dimensional since the $y_n$ are scalars.  
The input to the $l$-th layer is defined as~${\mathbf{a}_l = \mathbf{W}_l \mathbf{z}_{l-1} / \sqrt{V_{l-1}+1} }$,
where the factor~${1/\sqrt{V_{l-1} + 1}}$ keeps the scale of the input to each neuron independent
of the number of incoming connections.
The activation functions for each hidden layer are rectified linear units (ReLUs)
\citep{nair2010rectified}, i.e.,~${a(x) = \max(x,0)}$.


Let $\mathbf{y}$ be an $N$-dimensional vector with the targets~$y_n$
and~$\mathbf{X}$ be an ${N\times D}$ matrix of feature vectors~$\mathbf{x}_n$.
The likelihood for the network weights~$\mathcal{W}$ and the noise
precision~$\gamma$, with data~${\mathcal{D}=(\mathbf{X},\mathbf{y})}$ is then 
\begin{align}
p(\mathbf{y}\given\mathcal{W},\mathbf{X},\gamma) &= \prod_{n=1}^N \mathcal{N}(y_n\given f(\mathbf{x}_n;\mathcal{W}),\gamma^{-1})\,.\label{eq:likelihood}
\end{align}
To complete our probabilistic model, we specify a Gaussian prior distribution
for each entry in each of the weight matrices in $\mathcal{W}$. In particular,
\begin{align}
p(\mathcal{W}\given\lambda) &= \prod_{l=1}^L \prod_{i=1}^{V_l} \prod_{j=1}^{V_{l-1}+1} \mathcal{N}(w_{ij,l}\given0,\lambda^{-1})\,,\label{eq:prior_weights}
\end{align}
where $w_{ij,l}$ is the entry in the~$i$-th row and~$j$-th column
of~$\mathbf{W}_l$ and~$\lambda$ is a precision parameter. The hyper-prior
for~$\lambda$ is chosen to be a gamma distribution, i.e.,~${p(\lambda) = \DistGam(\lambda\given\alpha_0^\lambda,\beta_0^\lambda)}$ with shape~${\alpha^\lambda_0 = 6}$ and inverse scale~${\beta^\lambda_0 = 6}$.
The values chosen for $\alpha^\lambda_0$ and $\beta^\lambda_0$
make this equivalent to having observed~${\nu = 12}$  samples
from~$\mathcal{N}(0,\lambda^{-1})$ with empirical variance equal to 1. This
relatively low value for~$\nu$ compared to the large number $N$ of observed data points makes this prior weakly-informative. 
The prior for the noise precision~$\gamma$ is also gamma:~${p(\gamma) =
\DistGam(\gamma\given\alpha_0^{\gamma},\beta_0^{\gamma})}$.  We assume that the
$y_n$ have been normalized to have unit variance and, as above, we
fix~${\alpha^{\gamma}_0 = 6}$ and~${\beta^{\gamma}_0 = 6}$.

The posterior distribution for the parameters~$\mathcal{W}$,~$\gamma$
and~$\lambda$ can then be obtained by applying Bayes' rule:
\begin{align}
p(\mathcal{W},\gamma, \lambda\given\mathcal{D}) &=
\frac{p(\mathbf{y}\given\mathcal{W},\mathbf{X},\gamma)p(\mathcal{W}\given\lambda)p(\lambda)p(\gamma)}{p(\mathbf{y}\given\mathbf{X})}\,.\label{eq:exact_posterior}
\end{align}
where~$p(\mathbf{y}\given\mathbf{X})$ is a normalization constant. Given a new input
vector~$\mathbf{x}_\star$, we can then make predictions for its
output~$y_\star$ using the predictive distribution given by

\vspace{-0.6cm}
{\small
\begin{align}
p(y_\star\given\mathbf{x}_\star,\mathcal{D}) &= \int\!\! p(y_\star|\mathbf{x}_\star
\mathcal{W},\gamma) p(\mathcal{W},\gamma, \lambda|\mathcal{D})\,d\gamma\, d\lambda\, d\mathcal{W},\label{eq:predictive_distribution}
\end{align}
}where ${p(y_\star\given\mathbf{x}_\star,\mathcal{W},\gamma) =
\mathcal{N}(y_\star\given f(\mathbf{x}_\star), \gamma)}$. However, the exact
computation of  $p(\mathcal{W},\gamma, \lambda\given\mathcal{D})$ and
$p(y_\star\given\mathbf{x}_\star)$ is not tractable in most cases. Therefore, in
practice we have to resort to approximate inference methods. In the following
section we describe a technique for approximate Bayesian inference in NN models
that is both fast and also offers excellent predictive performance.

\section{Probabilistic backpropagation}\label{sec:method}

Backpropagation \cite{RumelhartHintonWIlliams1986} is by far the most common
method for training neural networks. This method operates in two phases to
compute a gradient of the loss in terms of the network weights. In the first
phase, the input features are propagated forward through the network to compute
the function output and thereby the loss associated with the current parameters.  In
the second phase, the derivatives of the training loss with respect to the
weights are propagated back from the output layer towards the input.
These derivatives are used to update the weights using, e.g.,
stochastic gradient descent with momentum.

In this section we describe a probabilistic alternative to the backpropagation
algorithm, which we call probabilistic backpropagation (PBP). PBP does not use
point estimates for the synaptic weights in the network. Instead, it uses a
collection of one-dimensional Gaussians, each one approximating the marginal
posterior distribution of a different weight. PBP also has two phases
equivalent to the ones of BP.  In the first phase, the input data is propagated
forward through the network.  However, since the weights are now random, the
activations produced in each layer are also random and result in (intractable)
distributions.  PBP sequentially approximates each of these distributions with
a collection of one-dimensional Gaussians that match their marginal means and
variances. At the end of this phase, PBP computes, instead of the prediction
error, the logarithm of the marginal probability of the target variable. In the
second phase, the gradients of this quantity with respect to the means and
variances of the approximate Gaussian posterior are propagated back using
reverse-mode differentiation as in classic backpropagation.  These derivatives
are finally used to update the means and variances of the posterior
approximation.

The update rule used by PBP is not the standard step in the direction of the
gradient of the loss made by the classic backpropagation algorithm.  PBP uses
the following property of Gaussian distributions \citep{minka2001family}. Let~$f(w)$ encode an arbitrary likelihood function for the single synaptic
weight~$w$ given some data and let our current beliefs regarding the scalar~$w$ be
captured by a distribution~${q(w)=\mathcal{N}(w\given m,v)}$.  After seeing the data,
our beliefs about $w$ are updated according to Bayes' rule:
\begin{align}
s(w) &= Z^{-1}f(w)\,\mathcal{N}(w\given m,v)\,,
\end{align}
where $Z$ is the normalization constant. The
updated beliefs~$s(w)$ usually have a complex form and need to be approximated
with a simpler distribution.  A common choice is to approximate this posterior
with a distribution that has the same form as~$q$.  In this case, the
parameters of the new Gaussian
beliefs~${q^\text{new}(w)=\mathcal{N}(w\given m^\text{new},v^\text{new})}$ that
minimize the the Kullback-Leibler (KL) divergence between~$s$ and~$q^\text{new}$ can
then be obtained as a function of~$m$, $v$ and the gradient of~$\log Z$ with
respect to these quantities, namely 
\begin{align}
m^\text{new} & = m + v \frac{\partial \log Z}{\partial m }\,,\label{eq:gaussian_update_m}\\
v^\text{new} & = v - v^2 \left[ \left(\frac{\partial \log Z}{\partial m } \right)^2 - 
2 \frac{\partial \log Z}{\partial v } \right]\,.\label{eq:gaussian_update_v}
\end{align}
See \cite{minka2001family}, equations 5.12 and 5.13.
These rules match moments between $q^\text{new}$ and $s$, guaranteeing that these two
distributions have the same mean and variance. 
These are the main update
equations used by PBP. The next section provides a detailed description of PBP,
presenting it as an \emph{assumed density filtering} (ADF) method
\cite{opper1998bayesian} that uses some of the improvements on ADF given by
expectation propagation \cite{minka2001family}.

\subsection{PBP as an assumed density filtering method}

Probabilistic backpropagation is an inference method that approximates the
exact posterior of a neural network~(\ref{eq:exact_posterior}) with a factored
distribution given by 
\begin{multline}
q(\mathcal{W},\gamma, \lambda) = \textstyle \left[ \prod_{l=1}^L\! \prod_{i=1}^{V_l}\! 
\prod_{j=1}^{V_{l\!-\!1}\!+\!1} \mathcal{N}(w_{ij,l}| m_{ij,l},v_{ij,l})\right]\\
 \times \DistGam(\gamma\given \alpha^\gamma, \beta^\gamma)
 \DistGam(\lambda\given \alpha^\lambda, \beta^\lambda)\,.\label{eq:posterior_approximation}
\end{multline}
The approximation parameters~$m_{ij,l}$,
$v_{ij,l}$,~$\alpha^\gamma$, $\beta^\gamma$, $\alpha^\lambda$ and~$\beta^\lambda$ are determined by applying an assumed density filtering method
\cite{opper1998bayesian,minka2001family} on the posterior
(\ref{eq:exact_posterior}). For this,~(\ref{eq:posterior_approximation}) is first
initialized to be uniform, that is,~${m_{ij,l}=0}$,~${v_{ij,l}=\infty}$,
${\alpha^\gamma=\alpha^\lambda=1}$ and~${\beta^\gamma=\beta^\lambda=0}$. After
this, PBP iterates over the factors in the numerator
of~(\ref{eq:exact_posterior}) and sequentially incorporates each of these
factors into the approximation in~(\ref{eq:posterior_approximation}). A general
description of this operation is given in the following paragraph.
Specific details on how to incorporate each type of factor in~(\ref{eq:exact_posterior}) are
given in the following sections.

There are two factors for the priors on~$\gamma$ and~$\lambda$, a total of~$\prod_{l=1}^L V_l (V_{l-1} + 1)$ factors for the prior on $\mathcal{W}$ given
by~(\ref{eq:prior_weights}) and finally,~$N$ factors for the
likelihood~(\ref{eq:likelihood}).  Let~$f(\mathcal{W},\lambda,\gamma)$ be one
of these factors. PBP incorporates~$f$ into the current posterior
approximation (\ref{eq:posterior_approximation}) by minimizing the
KL divergence between~${s(\mathcal{W},\gamma, \lambda) =
Z^{-1} f(\mathcal{W},\lambda,\gamma)q(\mathcal{W},\gamma, \lambda)}$
and~$q(\mathcal{W},\gamma, \lambda)$, with respect to the parameters of~$q$,
where~$Z$ normalizes~$s$ and the~$q$ used to construct~$s$ is kept constant
during the minimization of the KL divergence.  This makes the new $q$
approximate the product of the old $q$ and the factor $f$. The result
is an approximation to an exact online learning algorithm within the
Bayesian framework \cite{opper1998bayesian}.

\subsection{Incorporating the prior factors into $q$}

The first factors to be incorporated are the priors on~$\gamma$ and~$\lambda$. Since these factors have the same functional form
as~(\ref{eq:posterior_approximation}), the resulting update for~$q$ is
straightforward:~${\alpha^\gamma_\text{new} = \alpha_0^\gamma}$,
${\beta^\gamma_\text{new} = \beta_0^\gamma}$, ${\alpha^\lambda_\text{new} =
\alpha_0^\lambda}$ and ${\beta^\lambda_\text{new} = \beta_0^\lambda}$.

After this, we sequentially incorporate the factors
in~(\ref{eq:prior_weights}). In this case, the updates for~$m_{ij,l}$
and~$v_{ij,l}$ in~(\ref{eq:posterior_approximation}) are given
by~(\ref{eq:gaussian_update_m}) and~(\ref{eq:gaussian_update_v}). Similar
update rules can be obtained for the parameters~$\alpha^\lambda$
and~$\beta^\lambda$ in (\ref{eq:posterior_approximation}). In particular,
\begin{align}
\alpha^\lambda_\text{new} & = \left[ Z Z_2 Z_1^{-2} (\alpha^\lambda\!+\!1) / \alpha^\lambda - 1.0 \right]^{-1}\,,\label{eq:update_alpha}\\
\beta^\lambda_\text{new} & = \left[ Z_2 Z_1^{-1} (\alpha^\lambda\!+\!1) / \beta^\lambda - 
Z_1 Z^{-1} \alpha^\lambda / \beta^\lambda \right]^{-1}\,,\label{eq:update_beta}
\end{align}
where $Z$ is the normalizer of~$s$, that is, the product of the factor that is being incorporated and
$q$ and $Z_1$ and $Z_2$ are the values of this normalizer
when the parameter~$\alpha^\lambda$ in~$q$ is increased by
one and two units, respectively. The update rules~(\ref{eq:update_alpha})
and~(\ref{eq:update_beta}) do not exactly minimize the KL divergence since that
would require matching the sufficient statistics for~$\lambda$ in~$q$ and~$s$,
which does not have a closed form. Instead, the rules above match the first and
second moments of~$\lambda$, which also produces good results
\cite{minka2001family,cowell1996}. 
The derivation of~(\ref{eq:update_alpha}) and~(\ref{eq:update_beta}) can be
found in the supplementary material.
One difficulty when applying the update
rules just described is that the normalizer~$Z$ of~$s$ does not have a closed
form. Nevertheless, we can approximate~$Z$ using

\vspace{-0.50cm}
{\small
\hspace{-0.5cm}
\begin{align}
Z & = \textstyle \int \mathcal{N}(w_{ij,l}\given 0,\lambda^{-1}) q(\mathcal{W},\gamma, \lambda)\,d\mathcal{W}\,d\gamma\,d\lambda\nonumber \\
& = \textstyle \int \mathcal{N}(w_{ij,l}\given 0,\lambda^{-1}) \mathcal{N}(w_{ij,l}\given m_{ij,l},v_{ij,l})\nonumber\\
& \textstyle \quad\,\,\, \times \DistGam(\lambda\given \alpha^\lambda,\beta^\lambda)dw_{ij,l}\,d\lambda \nonumber\\
& = \textstyle \int \mathcal{T}(w_{ij,l}\given 0, \beta^\lambda/\alpha^\lambda, 2 \alpha^\lambda)\mathcal{N}(w_{ij,l}\given m_{ij,l},v_{ij,l})dw_{ij,l}\nonumber\\
& \approx \textstyle \int \mathcal{N}(w_{ij,l}\given 0, \beta^\lambda/(\alpha^\lambda - 1))
\mathcal{N}(w_{ij,l}\given m_{ij,l},v_{ij,l})dw_{ij,l}\nonumber\\
& = \textstyle \mathcal{N}(m_{ij,l}\given 0, \beta^\lambda/(\alpha^\lambda - 1) + v_{ij,l})\,.\label{eq:approximation_Z_prior}
\end{align}
}where $\mathcal{T}(\cdot\given \mu,\beta,\nu)$ denotes a Student's $t$ distribution
with mean $\mu$, variance parameter $\beta$ and degrees of freedom $\nu$.  In
the next-to-last line we have approximated the Student's $t$ density with a
Gaussian density that has the same mean and variance. 
Finally, $Z_1$, $Z_2$ and
the gradients of $\log Z$ with respect to $m_{ij,l}$ and $v_{ij,l}$ can be
similarly approximated by incorporating this approximation of $Z$ into their
expressions. By plugging in the resulting quantities
in~(\ref{eq:gaussian_update_m}),~(\ref{eq:gaussian_update_v}),~(\ref{eq:update_alpha})
and~(\ref{eq:update_beta}) we obtain the new parameter values for~$q$.

\subsection{Incorporating the likelihood factors into $q$}

After incorporating all the factors in~(\ref{eq:prior_weights}), PBP
sequentially incorporates the~$N$ factors for the
likelihood~(\ref{eq:likelihood}).  As before, updates for all the~$m_{ij,l}$
and~$v_{ij,l}$ in~(\ref{eq:posterior_approximation}) are given
by~(\ref{eq:gaussian_update_m}) and~(\ref{eq:gaussian_update_v}), respectively.
The updates for~$\alpha^\gamma$ and~$\beta^\gamma$
in~(\ref{eq:posterior_approximation}) are given by~(\ref{eq:update_alpha})
and~(\ref{eq:update_beta}), respectively. To compute all these updates we only
require~$Z$, the normalization constant of~$s$.  However, this is difficult to
compute, as it requires integration of each likelihood factor with respect to
the distribution of the network output, i.e.,~${z_L =
f(\mathbf{x}_n\given \mathcal{W})}$, when~${\mathcal{W}\sim q}$.
Let us assume that we have an approximating Gaussian with
mean~$m^{z_L}$ and variance~$v^{z_L}$ for the distribution of~$z_L$. We can then
approximate~$Z$ as

\vspace{-0.5cm}
{\small
\begin{align}
Z & = \textstyle \int \mathcal{N}(y_n\given f(\mathbf{x}_n\given \mathcal{W}),\gamma^{-1})q(\mathcal{W},\gamma,\lambda)\,d
\mathcal{W}\,d\gamma,d\lambda\nonumber\\
& \approx \textstyle \int \mathcal{N}(y_n| z_L,\gamma^{-1})\mathcal{N}(z_L| m^{z_L},v^{z_L})
\DistGam(\gamma\given \alpha^\gamma,\beta^\gamma)z_Ld\gamma\nonumber\\
& = \textstyle \int \mathcal{T}(y_n\given z_L,\beta^\gamma/\alpha^\gamma,2\alpha^\gamma)\mathcal{N}(z_L\given m^{z_L},v^{z_L})\,d z_L\nonumber\\
& \approx \textstyle \mathcal{N}(y_n\given m^{z_L},\beta^\gamma/(\alpha^\gamma-1)+ v^{z_L})\label{eq:approximation_Z_likelihood}
\end{align}
}where the first approximation in (\ref{eq:approximation_Z_likelihood}) assumes
that ${z_L = f(\mathbf{x}_i\given \mathcal{W})\sim\mathcal{N}(m^{z_L},v^{z_L})}$
when~$\mathcal{W}\sim q$ and the second approximates the
Student's $t$ density with a Gaussian density that has the same mean and
variance. An analysis of the error of this latter approximation can be found in the supplementary material.
This expression for~$Z$ can be substituted
into~(\ref{eq:gaussian_update_m}),~(\ref{eq:gaussian_update_v}),~(\ref{eq:update_alpha})
and~(\ref{eq:update_beta}) to obtain the new parameters for~$q$.

However, it remains to compute the mean and variance parameters~$m^{z_L}$
and~$v^{z_L}$ in~(\ref{eq:approximation_Z_likelihood}).  This is done by
propagating distributions forward through the network and, when necessary,
approximating each new distribution with a Gaussian.  For this, let us assume
that, when~$\mathcal{W}\sim q$, the output of the~${l-1}$
layer~$\mathbf{z}_{l-1}$ is a diagonal Gaussian with means and variances given
by the $V_{l-1}$-dimensional vectors~$\mathbf{m}^{\mathbf{z}_{l-1}}$
and~$\mathbf{v}^{\mathbf{z}_{l-1}}$, respectively. Furthermore,
let~$\mathbf{a}_l = \mathbf{W}_l \mathbf{z}_{l-1}/\sqrt{V_{l-1}+1}$, so that the marginal
means and variances of $\mathbf{a}_l$ (when $\mathcal{W}$ is distributed
as~$q$) are
\begin{align}
\mathbf{m}^{\mathbf{a}_l} & = \textstyle \mathbf{M}_l \mathbf{m}^{\mathbf{z}_{l-1}}/ \sqrt{V_{l-1}+1}\,,\label{eq:mean_a}\\
\mathbf{v}^{\mathbf{a}_l} & = \textstyle \left[(\mathbf{M}_l \!\circ \mathbf{M}_l) \mathbf{v}^{\mathbf{z}_{l\!-\!1}} + \mathbf{V}_l (\mathbf{m}^{\mathbf{z}_{l\!-\!1}}\!\circ \mathbf{m}^{\mathbf{z}_{l\!-\!1}})\right.\nonumber\\
& \quad + \left.  \mathbf{V}_l \mathbf{v}^{\mathbf{z}_{l-1}}\right] / (V_{l-1}+1)\label{eq:variance_a}
\end{align}
where $\mathbf{M}_l$ and $\mathbf{V}_l$ are ${V_l\times(V_{l-1}+1)}$ matrices whose
entries are given by~$m_{ij,l}$ and~$v_{ij,l}$, for~${i=1,\ldots,V_l}$
and~${j=1,\ldots,V_{l-1}+1}$, respectively, and~$\circ$ denotes the Hadamard
elementwise product. We again assume that the entries in~$\mathbf{a}_l$ are
independent Gaussian with means and variances given by the equations above. The
Central Limit Theorem states that~$\mathbf{a}_l$ is approximately Gaussian
when~$V_{l-1}$ is large \citep{Soudry2014}. Let ${\mathbf{b}_l = a(\mathbf{a}_l)}$,
where~$a$ is the rectifier linear activation function~${a(x) = \max(0,x)}$.
Then, the entries of~$\mathbf{b}_l$ are a mixture of a point mass at 0 (when
the rectifier is saturated) and a Gaussian truncated at 0 (when the rectifier
is in the linear regime). The mean and variance of the $i$-th entry of
$\mathbf{b}_l$ are then
\begin{align}
\textstyle m^{\mathbf{b}_l}_i & = \Phi(\alpha_i)  v_i' \,,\label{eq:mean_b}\\
\textstyle v^{\mathbf{b}_l}_i & = m^{\mathbf{b}_l}_i v_i' \Phi(-\alpha_i) +
\Phi(\alpha_i) v^{\mathbf{a}_l}_i (1\!-\!\gamma_i (\gamma_i\!+\!\alpha_i))\,,\label{eq:variance_b}
\end{align}
where
\begin{align*}
	v_i'\!&=\textstyle\!m^{\mathbf{a}_l}_i\! +\! \sqrt{v^{\mathbf{a}_l}_i} \gamma_i, &
\alpha_i \!&=\! \textstyle\frac{m^{\mathbf{a}_l}_i}{\sqrt{v^{\mathbf{a}_l}_i}},
& \gamma_i \!&=\! \textstyle\frac{\phi(-\alpha_i)}{\Phi(\alpha_i)},
\end{align*}
and~$\Phi$ and~$\phi$ are
respectively the CDF and the density function of a standard Gaussian. When
$\alpha_i$ is very large and negative the previous definition of $\gamma_i$ is
not numerically stable.  Instead, when ${\alpha_i < -30}$, we use the
approximation~${\gamma_i = -\alpha_i - \alpha_i^{-1} + 2 \alpha_i^{-3}}$ as
recommended by \citet{paquet2012}. The output of the $l$-th layer,
$\mathbf{z}_{l}$, is obtained by concatenating~$\mathbf{b}_l$ with the constant
1 for the bias. We can therefore approximate the distribution
of~$\mathbf{z}_{l}$ to be Gaussian with marginal means and variances
\begin{align}
\mathbf{m}^{\mathbf{z}_l} & = [ \mathbf{m}^{\mathbf{b}_l} \,;\, 1\, ]\,, &
\mathbf{v}^{\mathbf{z}_l} & = [ \mathbf{v}^{\mathbf{b}_l} \,;\, 0\, ]\,. \label{eq:mean_variance_next_layer}
\end{align}
These concatenated means and variances reflect the lack on
uncertainty in the ``bias unit''.  Finally, to compute the mean and variance
parameters~$m^{z_L}$ and~$v^{z_L}$ in~(\ref{eq:approximation_Z_likelihood}) we
initialize~$\mathbf{m}^{\mathbf{z}_0}$ to~$[ \mathbf{x}_i \, ; \, 1\, ]$
and~$\mathbf{v}^{\mathbf{z}_0}$ to~$\mathbf{0}$ and then
apply~(\ref{eq:mean_a}),~(\ref{eq:variance_a}),~(\ref{eq:mean_b}),~(\ref{eq:variance_b})
and~(\ref{eq:mean_variance_next_layer}) iteratively until we obtain~${m^{z_L}=
m_1^{\mathbf{a}_L}}$ and~${v^{z_L}= v_1^{\mathbf{a}_L}}$.  This resembles the
forward pass of the standard backpropagation algorithm.  With~$m^{z_L}$
and~$v^{z_L}$, we can evaluate the log of~$Z$ as given
by~(\ref{eq:approximation_Z_likelihood}) and the gradients of that quantity
that are required to apply rules~(\ref{eq:gaussian_update_m})
and~(\ref{eq:gaussian_update_v}).  This is similar to the reverse mode
differentiation used in backpropagation.  We provide a Theano-based \cite{bergstra2010} implementation of PBP at \url{http://jmhl.org/}, as well as a C version using the gradients given in the supplementary material.

\subsection{Expectation propagation}

Expectation propagation (EP) \cite{minka2001family} improves on assumed density
filtering by iteratively incorporating each factor multiple times. On each pass
over the list of factors, each factor is \emph{removed} from the current
posterior approximation, re-estimated, and then reincorporated.  Each iteration
improves the accuracy of the posterior approximation.  The disadvantage of EP
over ADF is that it needs to keep in memory all of the approximate factors, one
for each exact factor in the numerator of the posterior.  This is necessary,
because each factor must be able to be removed and updated.  With massive data
sets, the number of likelihoods will be very large and it is not possible to
store these factors in memory.  Instead, we incorporate these factors multiple
times, but without removing them from the current approximation. This is
equivalent to doing multiple ADF passes through the data, treating each
likelihood factor as a novel example.  A disadvantage of this approach is that
it can lead to underestimation of the variance parameters
in~(\ref{eq:posterior_approximation}) when too many passes are done over the
data. Nevertheless, PBP is geared toward larger data sets, where only a reduced
number of passes over the data (fewer than 100) are possible. Note that we can
afford to keep in memory an approximate factor for each exact factor in the
prior on the weights (\ref{eq:prior_weights}), since the number and size of
these approximate factors are small. We therefore do one full EP update of
these approximate factors for the prior after each ADF pass through the data.
Details for this operation can be found in the in the supplementary
material. These approximate factors could also be updated more frequently, for
example, each time we do an ADF pass through a small block of likelihood
factors.

\subsection{Implementation details}

After incorporating the factors in (\ref{eq:prior_weights}) for the first time,
we slightly perturb each mean parameter~$m_{ij,l}$
in~(\ref{eq:posterior_approximation}) from the original value of 0 to
be~$\epsilon_{ij,l}$, where~${\epsilon_{ij,l} \sim \mathcal{N}(0, 1 / (V_l+1))}$.
This is similar to the random initialization of weights in NNs that is usually
done before learning with backpropagation. This operation directs our 
inference method towards one of the multiple symmetric modes of the posterior.

Because the computation of $Z$ in (\ref{eq:approximation_Z_likelihood}) is
approximate, on rare occasions the variance parameters for some weights  in
(\ref{eq:posterior_approximation}) may be negative after incorporating one
likelihood factor. When this happens, we undo the update for those weights and
keep their previous mean and variance values. A similar operation is often done
in EP when negative variances arise in Gaussian approximate factors \cite{minka2001family}.

\newcommand{\ica}{\hspace{0.25cm}}

\renewcommand{\arraystretch}{0.94}

\begin{table*}
\centering
\resizebox{\textwidth}{!}{%
\begin{tabular}{lrrr@{$\pm$}l@{\ica}r@{$\pm$}l@{\ica}r@{$\pm$}l@{\ica}r@{$\pm$}l@{\ica}r@{$\pm$}l@{\ica}rrr}
& & & \multicolumn{6}{c}{Avg. Test RMSE and Std. Errors}
& \multicolumn{4}{c}{Avg. Test LL and Std. Errors}
& \multicolumn{3}{c}{Avg. Running Time in Secs}\\
\textbf{Dataset} & $N$ & $d$
    & \multicolumn{2}{c}{\bf{ VI }}&\multicolumn{2}{c}{\bf{ BP }}&\multicolumn{2}{c}{\bf{ PBP }}
    & \multicolumn{2}{c}{\bf{ VI }}&\multicolumn{2}{c}{\bf{ PBP }}
    & \multicolumn{1}{c}{\bf{ VI }} & \multicolumn{1}{c}{ \bf{BP} } & \multicolumn{1}{c}{ \bf{PBP}} \tabularnewline
\hline
Boston Housing & 506 & 13
   & 4.320&0.2914 & 3.228&0.1951 & \bf{3.014}&\bf{0.1800}
   & -2.903&0.071 & \bf{-2.574}&\bf{0.089}
   & 1035& 677& \bf{ 13 } \tabularnewline
Concrete Compression Strength & 1030 & 8
   & 7.128&0.1230 & 5.977&0.2207 & \bf{5.667}&\bf{0.0933}
   & -3.391&0.017 & \bf{-3.161}&\bf{0.019}
   & 1085& 758& \bf{ 24 } \tabularnewline
Energy Efficiency & 768 & 8
   & 2.646&0.0813 & \bf{1.098}&\bf{0.0738} & 1.804&0.0481
   &-2.391&0.029&\bf{-2.042}&\bf{0.019}
   & 2011& 675& \bf{ 19 } \tabularnewline
Kin8nm & 8192 & 8
   & 0.099&0.0009 & \bf{0.091}&\bf{0.0015} & 0.098&0.0007
   & \bf{0.897}&\bf{0.010} & 0.896&0.006
   & 5604& 2001& \bf{ 156 } \tabularnewline
Naval Propulsion & 11,934 & 16
   & 0.005&0.0005 & \bf{0.001}&\bf{0.0001} & 0.006&0.0000
   & \bf{3.734}&\bf{0.116} & 3.731&0.006
   & 8373& 2351& \bf{ 220 } \tabularnewline
Combined Cycle Power Plant & 9568 & 4
   & 4.327&0.0352 & 4.182&0.0402 & \bf{4.124}&\bf{0.0345}
   & -2.890&0.010 & \bf{-2.837}&\bf{0.009}
   & 2955& 2114& \bf{ 178 } \tabularnewline
Protein Structure & 45,730 & 9 
   & 4.842&0.0305 & \bf{4.539}&\bf{0.0288} & 4.732&0.0130
   & -2.992&0.006 & \bf{-2.973}&\bf{0.003}
   & 7691& 4831& \bf{ 485 } \tabularnewline
Wine Quality Red & 1599 & 11
   & 0.646&0.0081 & 0.645&0.0098 & \bf{0.635}&\bf{0.0079}
   & -0.980&0.013 & \bf{-0.968}&\bf{0.014}
   & 1195& 917& \bf{ 50 } \tabularnewline
Yacht Hydrodynamics & 308 & 6
   & 6.887&0.6749 & 1.182&0.1645 & \bf{1.015}&\bf{0.0542}
   & -3.439&0.163 & \bf{-1.634}&\bf{0.016}
   & 954& 626& \bf{ 12 } \tabularnewline
Year Prediction MSD & 515,345 & 90
   & 9.034&   NA & 8.932 &   NA & \bf{8.879} & \bf{   NA}
   & -3.622&  NA & \bf{-3.603}&\bf{  NA}
   & 142,077& 65,131& \bf{ 6119 }\tabularnewline
\hline
\end{tabular}
}
\caption{Characteristics of the analyzed data sets, average test performance in RMSE and log likelihood, and average running time.}
\label{tab:results}
\vspace{-0.4cm}
\end{table*}

\section{Related Work}\label{sec:related_work}

The gold standard method for Bayesian learning in neural networks is Hamilton
Monte Carlo (HMC) \citep{neal1995bayesian}.  However, this is a batch method
that can perform poorly on large data sets. HMC also requires problem-specific
tuning parameters such as the length and number of integration
steps.  One alternative to MCMC inference in neural networks is the Laplace
approximation \citep{mackay1992practical}.  However, the Laplace approximation
requires computation of the inverse Hessian of the log likelihood, which can be
infeasible to compute for large networks. Diagonal approximations
to the Hessian are possible, but performance can deteriorate considerably.  One
alternative approach based on EP is described by \citet{jylanki2014expectation}.
This is a batch method that is not expected to scale to large
data sets and, unlike PBP, it requires numerical quadrature.
Jylanki keeps in memory several approximate factors for each data point, which
is not feasible in large scale settings.  Furthermore, by using latent variables,
this method breaks each likelihood factor into additional sub-factors that are
incorporated into the posterior approximation in multiple disconnected steps.
PBP incorporates each likelihood factor in a single step,
which is expected to be more accurate.

A scalable variational inference (VI) method for neural networks is described
by \citet{Graves2011}. This method maximizes a lower bound on the marginal
likelihood of the NN. The computation of this bound requires computing the
expectation of the log of the numerator of the exact
posterior~(\ref{eq:exact_posterior}) under a factorized Gaussian approximation.
This is intractable in general, and so \citet{Graves2011} proposes a Monte
Carlo approximation to compute the lower bound, which is then optimized using a
second approximation for stochastic gradient descent (SGD).  While SGD is a
common approach to optimization of neural networks, the initial approximation
leads to inefficient use of data.  As a result, the VI approach tends to
generate poor solutions for larger data sets over which only a few passes are
possible.

The technique that is most closely related to PBP is the
expectation-backpropagation (EBP) method described by \citet{Soudry2014}, which
proposes an online EP technique for neural networks with \emph{sign} activation
functions and binary weights, with an extension to continuous weights. As with
PBP, EBP also includes a forward propagation of probabilities followed by a
backward propagation of gradients. However, there are three important
contributions of PBP with respect to EBP. First,  EBP can only model data with
binary targets and cannot be applied when the~$y_n$ are continuous (as in
regression), while PBP assumes continuous $y_n$ and can be extended to binary
targets using the same method as in EBP. Second, and more importantly, EBP with
continuous weights only updates the mean parameters of the Gaussian posterior
approximations.  In particular, the EBP update operation for each Gaussian
approximation includes only equation~(\ref{eq:gaussian_update_m}) and does not
perform the corresponding update for the variance given
by~(\ref{eq:gaussian_update_v}).  Therefore, EBP cannot produce accurate
uncertainty estimates, as it keeps the posterior variances constant during the
learning process. Note also that the ``learning rate''
in~(\ref{eq:gaussian_update_m}) is the variance of the Gaussian approximation.
In effect, by not updating the variances, EBP makes inefficient updates for the
means.  Finally, unlike probabilistic backpropagation, EBP does not learn the
hyperparameter for the prior variance $\lambda^{-1}$.  Instead, EBP keeps
$\lambda^{-1}$ fixed to a large initial value.

\section{Experiments}\label{sec:experiments}

We evaluate PBP in regression experiments with publicly available data sets and
neural networks with one hidden layer.  In PBP we make probabilistic
predictions for the target variables by using
(\ref{eq:approximation_Z_likelihood}), which
approximates~(\ref{eq:predictive_distribution}).

\begin{figure}[t]
\centering
\includegraphics[width=\columnwidth]{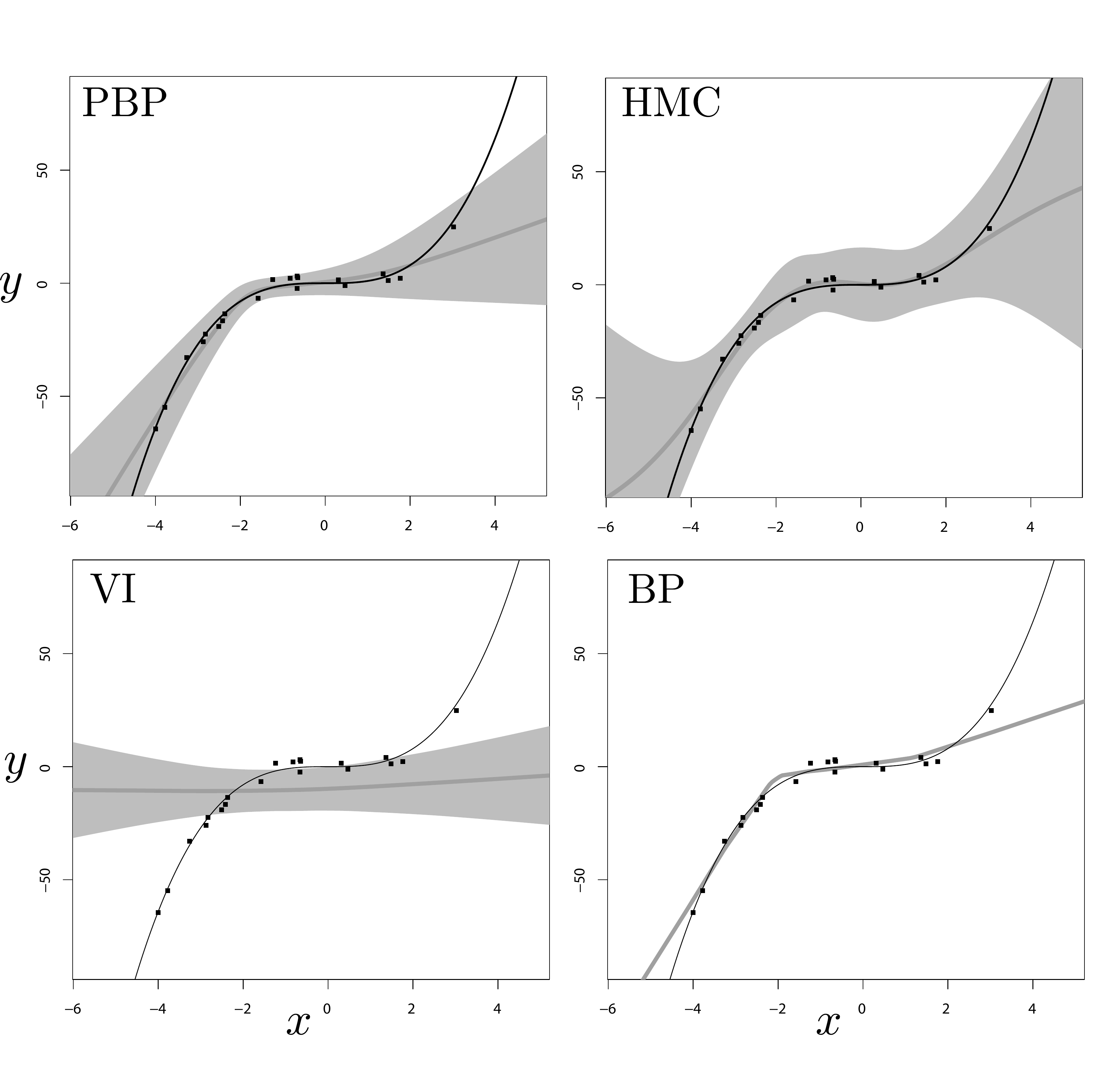}
\caption{Predictions made by each method on the toy data set.  The noisy
observations are shown as black dots, the true data generating function is
displayed as a black continuous line and the mean predictions are shown as a
dark gray line.  Credible intervals corresponding to $\pm3$ standard deviations
from the mean are shown as a light gray shaded area.
}
\label{fig:toy_dataset}
\end{figure}

\subsection{Predictive performance}

We first evaluate the predictive accuracy of PBP. Table~\ref{tab:results}
lists the analyzed data sets and shows summary statistics.
We use neural networks with 50 hidden units in all cases except in
the two largest ones, i.e., \emph{Year Prediction MSD} and \emph{Protein
Structure}, where we use 100 hidden units. We compare PBP with the variational
inference (VI) approach described in Section~\ref{sec:related_work} and with
standard stochastic gradient descent via backpropagation (BP).  These
methods were coded using Theano \citep{bergstra2010}.

The different methods, PBP, VI and BP, were run by performing 40 passes over
the available training data, updating the model parameters after seeing each
data point.  The data sets are split into random training and test sets with
90\% and 10\% of the data, respectively. This splitting process is repeated 20
times and the average test performance of each method is reported. In the two
largest data sets, \emph{Year Prediction MSD} and \emph{Protein Structure}, we
do the train-test splitting only one and five times respectively. The data sets
are normalized so that the input features and the targets have zero
mean and unit variance in the training set. The normalization on the targets
is removed for prediction.

\begin{figure*}
\centering
\includegraphics[width=0.85\textwidth]{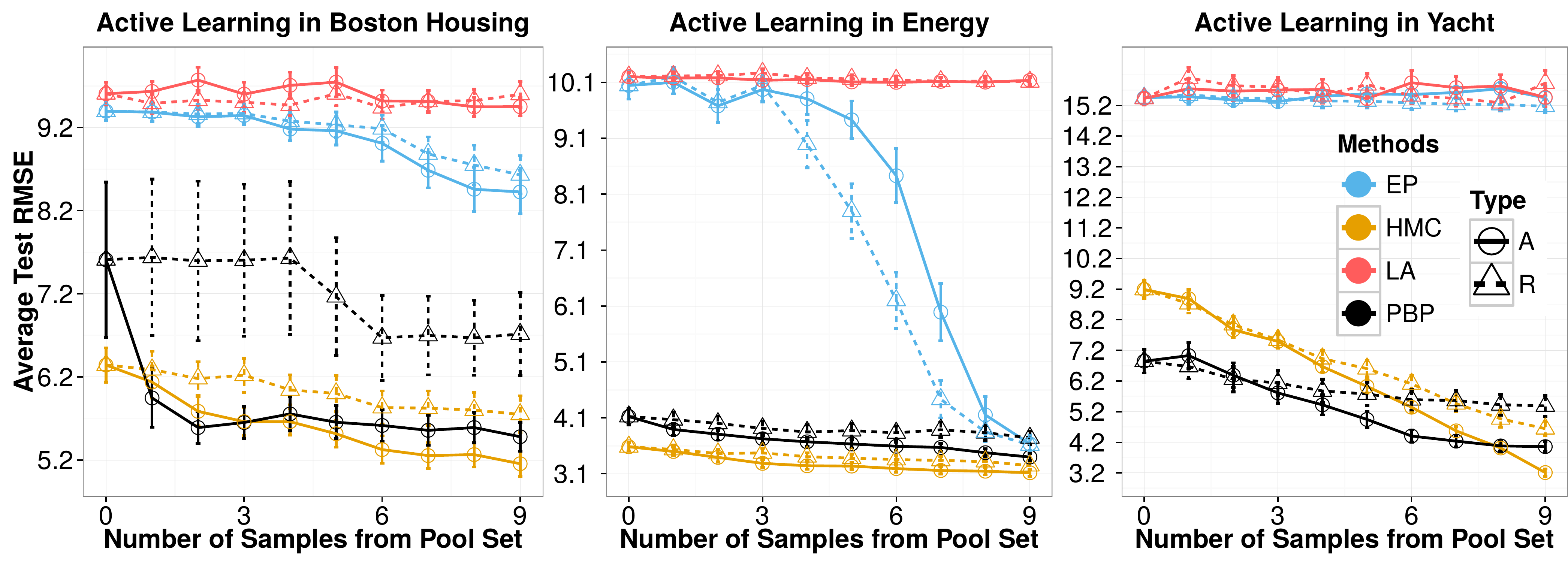}
\caption{Average test RMSE and standard errors in the active learning experiments with Boston Housing, Yacht and Energy data sets.}
\label{fig:active_learning}
\end{figure*}

BP and VI have several hyperparameters that have to be optimally adjusted to
the data. These are learning rate and momentum in BP and VI and weight decay in
BP.  We select these hyperparameter values by maximizing the predictive
performance of each method on a validation set with 20\% of the training data.
For this task we use Bayesian optimization (BO) techniques
\cite{snoek2012practical}.  In particular, for each train-test split of the
data, we use BO to sequentially evaluate the validation performance of
30 different hyperparameter configurations. After that, an optimal value for
the hyperparameters is selected and used to fit the model on the training set.

Table \ref{tab:results} shows the average test root mean squared error
(RMSE) for each method. On each data set, the results of the best method are
shown in bold.  Overall, PBP and BP perform best, with PBP obtaining the best results in
6 out of 10 data sets.  Unlike BP, PBP automatically adjusts its
hyperparameters and does not require an expensive BO search. VI performs rather
poorly, evidently due to the use of two stochastic approximations. First, VI
approximates the lower bound on the model evidence by sampling from the
variational approximation and second, VI further approximates that bound by
subsampling the data. BP and PBP only perform the second type of approximation.

Table \ref{tab:results} also shows the average test log-likelihood for VI and
PBP, and average running time for each method, in seconds. PBP is considerably better than VI, which performs rather poorly. BP and VI are very slow since they have to be re-run 30 times
to search for their optimal hyperparameter values. The BO search in these
methods also has a considerable overhead in the smallest data sets.  PBP is the
fastest method since it does not have to select any hyperparameter values and
is run only once.

\subsection{Multiple hidden layers}

A comparison of the test RMSE obtained by PBP and BP in neural networks with up
to 4 hidden layers can be found in the supplementary material.  The
experimental protocol in these experiments is the same as before. We use
networks with 50 units in each hidden layer, except in the datasets \emph{Year} and
\emph{Protein}, where we use 100. These results are similar to those
shown in Table \ref{tab:results}, with PBP obtaining usually the best
results with 2 hidden layers.

\subsection{Toy data set}

We further evaluate the predictive distribution obtained by PBP in a toy
data set generated by sampling 20 inputs~$x$ uniformly at random in the
interval~$[-4,4]$.  For each value of~$x$ obtained, the corresponding target
$y$ is computed as~${y = x^3 + \epsilon_n}$, where~${\epsilon_n \sim
\mathcal{N}(0, 9)}$. We fitted a neural network with one layer and 100 hidden
units to these data using PBP. We compare PBP with VI and BP, using 40 training
epochs in all these methods. We also compare with a ground truth generated by
Hamiltonian Monte Carlo (HMC).  HMC is implemented by modifying the MCMCstuff
Matlab toolbox \cite{vanhatalo2006mcmc} to include rectified linear activation
functions.  We run HMC by drawing 30,000 samples from the posterior
distribution.

Figure \ref{fig:toy_dataset} shows the predictions generated by each method.
PBP and BP are much closer to the ground truth HMC than VI. Furthermore, BP and
PBP perform similarly, even though PBP automatically adjusts its
hyperparameters while BP has to use BO methods for this task.

\begin{table*}
\centering
\vspace{-0.0cm}
\resizebox{\textwidth}{!}{%
\begin{tabular}{l@{\ica}r@{$\pm$}l@{\ica}r@{$\pm$}l@{\ica}r@{$\pm$}l@{\ica}r@{$\pm$}l@{\ica}r@{$\pm$}l@{\ica}r@{$\pm$}l@{\ica}r@{$\pm$}l@{\ica}r@{$\pm$}l@{\ica}}
\bf{Dataset}&\multicolumn{2}{c}{\bf{ LA-R }}&\multicolumn{2}{c}{\bf{ LA-A }}&\multicolumn{2}{c}{\bf{ EP-R }}&\multicolumn{2}{c}{\bf{ EP-A }}&\multicolumn{2}{c}{\bf{ PBP-R }}&\multicolumn{2}{c}{\bf{ PBP-A }}&\multicolumn{2}{c}{\bf{ HMC-R }}&\multicolumn{2}{c}{\bf{ HMC-A }}\\
\hline
Boston&9.600&0.154&9.452&0.111&8.632&0.231&8.426&0.264&6.716&0.500&5.480&0.175&5.750&0.222&\bf{5.156}&\bf{0.150}\\
Concrete&16.889&0.182&16.938&0.173&16.767&0.174&16.897&0.151&12.417&0.392&11.894&0.254&\bf{10.564}&\bf{0.198}&11.484&0.191\\
Energy&10.110&0.075&10.135&0.070&3.616&0.101&3.634&0.159&3.743&0.121&3.399&0.064&3.246&0.067&\bf{3.118}&\bf{0.062}\\
Kin8nm&0.271&0.003&0.270&0.002&0.272&0.002&0.271&0.002&0.259&0.006&0.254&0.005&0.226&0.004&\bf{0.223}&\bf{0.003}\\
Naval&0.015&0.000&0.015&0.000&0.015&0.000&0.015&0.000&0.015&0.000&0.016&0.000&0.013&0.000&\bf{0.012}&\bf{0.000}\\
Power Plant&17.195&0.120&17.306&0.149&8.234&0.831&6.251&0.599&5.312&0.108&5.068&0.082&5.229&0.097&\bf{4.800}&\bf{0.074}\\
Protein&6.165&0.073&6.227&0.088&6.118&0.074&6.151&0.077&6.133&0.141&5.903&0.127&\bf{5.613}&\bf{0.089}&5.727&0.090\\
Wine&0.843&0.011&0.829&0.010&0.836&0.010&0.832&0.009&0.945&0.044&0.809&0.011&\bf{0.740}&\bf{0.011}&0.749&0.010\\
Yacht&15.926&0.409&15.463&0.310&15.173&0.214&15.442&0.390&5.388&0.339&4.058&0.158&4.644&0.237&\bf{3.211}&\bf{0.120}\\
\hline
\end{tabular}
}
\caption{ Average test RMSE and standard errors in active learning. }\label{tab:active_learning}
\vspace{-0.40cm}
\end{table*}

\subsection{Active learning}

We performed another series of experiments to evaluate the accuracy of the
estimates of the posterior variance on the weights produced by PBP. For this,
we use an active learning scenario \citep{Settles2009} since in this type of
problems it is necessary to produce accurate estimates of uncertainty for
obtaining good performance.

In these experiments, we used a neural network with a single hidden layer and
ten hidden units.  We split each data set into training and test sets with 20
and 100 data instances, respectively, and pool sets with all the remaining data. PBP
is fitted using the training data and then, its performance is evaluated on the
test data. After this, one data point is collected from the pool set and then
moved into the training set. The process repeats until 9 of these active
additions to the training set have been completed, that is, until we have
performed 10 evaluations on the test set.  The entire process, including the
random data set splitting, is repeated 40 times.   The pool data is initially
lacking the target variables and these become available only once the data is
moved to the training set. As before, we run PBP for 40 epochs.

We compare PBP with a ground truth obtained by a HMC method in which we draw
500 samples from the posterior. We also compare with the batch EP algorithm for
neural networks described by \citet{jylanki2014expectation}.  This method uses
nonlinear activation functions given by the standard Gaussian CDF.  We further
compare with the Laplace approximation (LA) of \citet{mackay1992practical}
using the neural network toolbox from Matlab with tanh nonlinearities.  In LA
we approximate the Hessian of the unnormalized posterior distribution with the
Levenberg-Marquardt approximation and assume a diagonal Hessian matrix. This
allows LA to scale to large data sets and larger networks. We compare
two versions of PBP, HMC, EP and LA. One in which the data from
the pool set is collected actively (PBP-A, HMC-A, EP-A and LA-A) and another
one in which the pool data is collected uniformly at random (PBP-R, HMC-R, EP-R
and LA-R). We re-trained from scratch all the methods after each new addition
to the training set from the pool set.

To actively collect data from the pool set we follow the information-based
approach described by \citet{Mackay1992}. The goal is to maximize the expected
reduction in posterior entropy that is produced by adding data to the
training set.  This implies choosing the~$\mathbf{x}$ that maximizes
\begin{align}
\text{H}[\mathcal{W},\gamma,\lambda\given \mathcal{D}] - 
\mathbb{E}_{y\given \mathbf{x},\mathcal{D}}\text{H}[\mathcal{W},\gamma,\lambda\given \mathcal{D}\cup\{\mathbf{x},y\}]\,,\label{eq:acquisition_function}
\end{align}
where $\text{H}[\cdot]$ is the differential entropy.  Following \citet{houlsby2012collaborative}, we can rewrite~(\ref{eq:acquisition_function}) by swapping the roles of~$y$ and the model parameters~$\mathcal{W},\gamma,\lambda$. We finally obtain
\begin{align}
\text{H}[y\given \mathbf{x},\mathcal{D}] - 
\mathbb{E}_{\mathcal{W},\gamma,\lambda\given \mathcal{D}}\text{H}[y\given \mathcal{W},\gamma,\lambda,\mathbf{x}]\,.\label{eq:new_acquisition_function}
\end{align}
Since the last term in (\ref{eq:new_acquisition_function}) is constant, we
select the~$\mathbf{x}$ that maximizes the entropy of the predictive
distribution~$p(y\given \mathbf{x},\mathcal{D})$. Therefore, all the methods select
the next~$\mathbf{x}$ with highest predictive variance.

Table \ref{tab:active_learning} shows the average test RMSE for each method at
the end of the data collection process. These results show that the active
learning approach HMC-A is significantly better than the random approach HMC-R
in the data sets Boston, Energy, Power Plant and Yacht.  In these data sets we
also see a significant improvement of PBP-A over PBP-R. This indicates that PBP
produces useful estimates of posterior variance. In these
experiments PBP is usually better than EP and LA. LA performs poorly
because it cannot correctly select the hyperparameters $\lambda$ and $\gamma$,
due to the diagonal Hessian approximation, as also observed by
\citet{mackay1992bayesian}. PBP does not have this problem.

Finally, Figure \ref{fig:active_learning} shows the evolution of the average
test RMSE for each method during the data collection process in the problems
Boston, Yacht and Energy. These plots indicate that the improvements of PBP-A
over PBP-R are similar to those of HMC-A over HMC-R. Furthermore,  we can see
that the active learning strategy does not work as well in EP and LA as it does
in PBP and HMC.


\section{Conclusions and future work}\label{sec:conclusions}

We have presented probabilistic backpropagation (PBP), a new algorithm for
scalable Bayesian learning of neural networks.  PBP uses a product of Gaussians
to approximate the posterior over weights.  The parameters of these Gaussians
are updated in a two stage process similar to the one used by the
backpropagation algorithm.  First, probabilities are propagated forward through
the network to obtain the marginal likelihood and second, the gradients of this
quantity with respect to the Gaussian parameters are propagated backwards.
These gradients are finally used to update the parameters of the approximation
to the posterior distribution. Experiments on ten datasets show that PBP makes
accurate predictions.  Furthermore, we also show that PBP produces useful
estimates of the posterior variance on the network weights. In summary, PBP is
a fast method with state-of-the-art performance for Bayesian learning of neural
networks.  As future work we plan to address multi-label and
multi-class problems. We will also make PBP use mini-batches and
output estimates of the model evidence


\subsection*{Acknowledgements} 

The authors thank Edward O. Pyzer-Knapp for helpful discussions.
Jos\'e Miguel Hern\'andez-Lobato acknowledges support from the Rafael del Pino Foundation.  This work was partially funded by NSF IIS-1421780.



\section*{Supplementary material}

# Probabilistic Backpropagation for Scalable Learning of Bayesian Neural Networks
## *Supplementary Material*

José Miguel Hernández-Lobato, Harvard University, USA
Ryan P. Adams, Harvard University, USA

## 1 Derivation of the gradients

In this section we derive the gradient of the logarithm of the marginal likelihood, that is, $\log Z$, with respect to the means and variances of the network weights in the Gaussian approximation $q$. In traditional backpropagation we have, for each neuron $j$, one variable $\delta_j$ containing the gradient of the network error with respect to the input or activation for neuron $j$. In PBP, the corresponding algorithm is very similar, with the difference that we now have two variables for each neuron $j$ instead of only one. We have one variable $\delta_j^m$ that contains the gradient of $\log Z$ with respect to the mean of the activation for neuron $j$. Additionally, there is another variable $\delta_j^v$ that contains the gradient of $\log Z$ with respect to the variance of the activation for neuron $j$.

The mean and variance of the output of unit $j$ are defined as $m_j^z$ and $v_j^z$, respectively. The mean and variance of the activation or input for unit $j$ are defined as $m_j^a$ and $v_j^a$, respectively. We have that, becauseof the ReLU activation function,

$$m_j^z = \Phi(\alpha_j)\left[m_j^a + \sqrt{v_j^a}\gamma_j\right], \tag{1}$$

$$v_j^z = m_j^z \left[m_j^a + \sqrt{v_j^a}\gamma_j\right]\Phi(-\alpha_j) + \Phi(\alpha_j)v_j^a(1 - \gamma_j^2 - \gamma_j\alpha_j), \tag{2}$$

where $\gamma_j = \phi(\alpha_j)/\Phi(\alpha_j)$, $\alpha_j = m_j^a/\sqrt{v_j^a}$ and $\phi$ and $\Phi$ denote the standard Gaussian pdf and cdf. For the single neuron in the last layer we have that $m_j^z = m_j^a$ and $v_j^z = v_j^a$.

We have that $m_j^a$ and $v_j^a$ are given by

$$m_j^a = \frac{1}{\sqrt{|I(j)|}} \sum_{i \in I(j)} m_i^z m_{j,i}^w, \tag{3}$$

$$v_j^a = \frac{1}{|I(j)|} \sum_{i \in I(j)} \left\{ [m_i^z]^2 v_{j,i}^w + [v_i^z][m_{j,i}^w]^2 + v_i^z v_{j,i}^w \right\}, \tag{4}$$

where $I(j)$ is the set of neurons whose output is the input to neuron $j$, $m_{i,j}^w$ and $v_{i,j}^w$ are the mean and variances of the weight connecting neurons $i$ and $j$. Therefore,

$$\frac{\partial m_j^a}{\partial m_i^z} = \frac{1}{\sqrt{|I(j)|}} m_{j,i}^w, \qquad \frac{\partial m_j^a}{\partial v_i^z} = 0, \tag{5}$$

$$\frac{\partial v_j^a}{\partial m_i^z} = \frac{2 m_i^z v_{j,i}^w}{|I(j)|}, \qquad \frac{\partial v_j^a}{\partial v_i^z} = \frac{[m_{j,i}^w]^2 + v_{j,i}^w}{|I(j)|}, \tag{6}$$

and

$$\frac{\partial m_i^a}{\partial m_{i,j}^w} = \frac{m_j^z}{\sqrt{|I(i)|}}, \qquad \frac{\partial m_i^a}{\partial v_{i,j}^w} = 0, \tag{7}$$

$$\frac{\partial v_i^a}{\partial m_{i,j}^w} = \frac{2 v_j^z m_{i,j}^w}{|I(i)|}, \qquad \frac{\partial v_i^a}{\partial v_{i,j}^w} = \frac{[m_j^z]^2 + v_j^z}{|I(i)|}, \tag{8}$$

We now compute the gradient of $\gamma_j$ and $\alpha_j$ with respect to $m_j^a$ and $v_j^a$:

$$\frac{\partial \alpha_j}{\partial m_j^a} = \frac{1}{\sqrt{v_j^a}}, \qquad \frac{\partial \alpha_j}{\partial v_j^a} = \frac{m_j^a}{2 v_j^a \sqrt{v_j^a}}, \tag{9}$$

$$\frac{\partial \gamma_j}{\partial m_j^a} = -\left[\gamma_j \alpha_j + \gamma_j^2\right] \frac{\partial \alpha_j}{\partial m_j^a}, \qquad \frac{\partial \gamma_j}{\partial v_j^a} = -\left[\gamma_j \alpha_j + \gamma_j^2\right] \frac{\partial \alpha_j}{\partial v_j^a}. \tag{10}$$

Then we obtain

$$\frac{\partial m_j^z}{\partial m_j^a} = \frac{\partial \alpha_j}{\partial m_j^a} \phi(\alpha_j) \left[m_j^a + \sqrt{v_j^a}\gamma_j\right] + \Phi(\alpha_j)\left[1 + \sqrt{v_j^a}\frac{\partial \gamma_j}{\partial m_j^a}\right], \tag{11}$$

$$\frac{\partial m_j^z}{\partial v_j^a} = \frac{\partial \alpha_j}{\partial v_j^a} \phi(\alpha_j) \left[m_j^a + \sqrt{v_j^a}\gamma_j\right] + \Phi(\alpha_j)\left[\frac{\gamma_j}{2\sqrt{v_j^a}} + \sqrt{v_j^a}\frac{\partial \gamma_j}{\partial v_j^a}\right], \tag{12}$$

$$\frac{\partial v_j^z}{\partial m_j^a} = \frac{\partial m_j^z}{\partial m_j^a}\left[m_j^a + \sqrt{v_j^a}\gamma_j\right]\Phi(-\alpha_j) + m_j^z\left\{\left[1 + \sqrt{v_j^a}\frac{\partial \gamma_j}{\partial m_j^a}\right]\Phi(-\alpha_j) - \left[m_j^a + \sqrt{v_j^a}\gamma_j\right]\phi(\alpha_j)\frac{\partial \alpha_j}{\partial m_j^a}\right\} +$$

$$\phi(\alpha_j)\frac{\partial \alpha_j}{\partial m_j^a} v_j^a (1 - \gamma_j^2 - \gamma_j\alpha_j) - \Phi(\alpha_j)v_j^a\left\{2\gamma_j\frac{\partial \gamma_j}{\partial m_j^a} + \frac{\partial \gamma_j}{\partial m_j^a}\alpha_j + \gamma_j\frac{\partial \alpha_j}{\partial m_j^a}\right\}, \tag{13}$$

$$\frac{\partial v_j^z}{\partial v_j^a} = \frac{\partial m_j^z}{\partial v_j^a}\left[m_j^a + \sqrt{v_j^a}\gamma_j\right]\Phi(-\alpha_j) +$$

$$m_j^z \left\{ \left[\frac{1}{2\sqrt{v_j^a}}\gamma_j + \frac{\partial \gamma_j}{\partial v_j^a}\sqrt{v_j^a}\right]\Phi(-\alpha_j) - \left[m_j^a + \sqrt{v_j^a}\gamma_j\right]\phi(\alpha_j)\frac{\partial \alpha_j}{\partial v_j^a}\right\} +$$

$$\phi(\alpha_j)\frac{\partial \alpha_j}{\partial v_j^a}v_j^a(1 - \gamma_j^2 - \gamma_j\alpha_j) +$$

$$\Phi(\alpha_j)\left\{(1 - \gamma_j^2 - \gamma_j\alpha_j) + v_j^a\left\{-2\gamma_j\frac{\partial \gamma_j}{\partial v_j^a} - \frac{\partial \gamma_j}{\partial v_j^a}\alpha_j - \gamma_j\frac{\partial \alpha_j}{\partial v_j^a})\right\}\right\}. \tag{14}$$

We now define the variables $\delta_j^m$ and $\delta_j^v$ to be

$$\delta_j^m = \frac{\partial \log Z}{\partial m_j^a} = \sum_{k \in O(j)}\left\{\frac{\partial \log Z}{\partial m_k^a}\frac{\partial m_k^a}{\partial m_j^a} + \frac{\partial \log Z}{\partial v_k^a}\frac{\partial v_k^a}{\partial m_j^a}\right\}, \tag{15}$$

$$\delta_j^v = \frac{\partial \log Z}{\partial v_j^a} = \sum_{k \in O(j)}\left\{\frac{\partial \log Z}{\partial m_k^a}\frac{\partial m_k^a}{\partial v_j^a} + \frac{\partial \log Z}{\partial v_k^a}\frac{\partial v_k^a}{\partial v_j^a}\right\}, \tag{16}$$

where the sum is over each neuron $k$ to which neuron $j$ sends signals. The above rules can be recursively written as follows:

$$\delta_j^m = \frac{\partial \log Z}{\partial m_j^a} = \sum_{k \in O(j)}\left\{\delta_k^m\frac{\partial m_k^a}{\partial m_j^a} + \delta_k^v\frac{\partial v_k^a}{\partial m_j^a}\right\}, \tag{17}$$

$$\delta_j^v = \frac{\partial \log Z}{\partial v_j^a} = \sum_{k \in O(j)}\left\{\delta_k^m\frac{\partial m_k^a}{\partial v_j^a} + \delta_k^v\frac{\partial v_k^a}{\partial v_j^a}\right\}, \tag{18}$$

We can then write the required terms $\frac{\partial m_k^a}{\partial m_j^a}$, $\frac{\partial v_k^a}{\partial m_j^a}$, $\frac{\partial m_k^a}{\partial v_j^a}$ and $\frac{\partial v_k^a}{\partial v_j^a}$ as follows:

$$\frac{\partial m_k^a}{\partial m_j^a} = \frac{\partial m_k^a}{\partial m_j^z}\frac{\partial m_j^z}{\partial m_j^a} + \frac{\partial m_k^a}{\partial v_j^z}\frac{\partial v_j^z}{\partial m_j^a}, \qquad \frac{\partial v_k^a}{\partial m_j^a} = \frac{\partial v_k^a}{\partial m_j^z}\frac{\partial m_j^z}{\partial m_j^a} + \frac{\partial v_k^a}{\partial v_j^z}\frac{\partial v_j^z}{\partial m_j^a}, \tag{19}$$

$$\frac{\partial m_k^a}{\partial v_j^a} = \frac{\partial m_k^a}{\partial m_j^z}\frac{\partial m_j^z}{\partial v_j^a} + \frac{\partial m_k^a}{\partial v_j^z}\frac{\partial v_j^z}{\partial v_j^a}, \qquad \frac{\partial v_k^a}{\partial v_j^a} = \frac{\partial v_k^a}{\partial m_j^z}\frac{\partial m_j^z}{\partial v_j^a} + \frac{\partial v_k^a}{\partial v_j^z}\frac{\partial v_j^z}{\partial v_j^a}. \tag{20}$$

Table 1: Average Test RMSE in the experiments with deep neural networks.

| Dataset | $BP_1$ | $BP_2$ | $BP_3$ | $BP_4$ | $PBP_1$ | $PBP_2$ | $PBP_3$ | $PBP_4$ |
|---|---|---|---|---|---|---|---|---|
| Boston | 3.228±0.1951 | 3.185±0.2365 | 3.019±0.1848 | 2.874±0.1570 | 3.014±0.1800 | **2.795±0.1590** | 2.938±0.1645 | 3.088±0.1519 |
| Concrete | 5.977±0.2207 | 5.396±0.1273 | 5.568±0.1271 | 5.530±0.1390 | 5.667±0.0933 | **5.241±0.1164** | 5.732±0.1075 | 5.956±0.1597 |
| Energy | 1.185±0.1242 | 0.676±0.0367 | **0.628±0.0278** | 0.667±0.0321 | 1.804±0.0481 | 0.903±0.0482 | 1.237±0.0592 | 1.176±0.0552 |
| Kin8nm | 0.091±0.0015 | 0.073±0.0009 | 0.071±0.0006 | 0.071±0.0009 | 0.098±0.0007 | **0.071±0.0005** | 0.073±0.0007 | 0.075±0.0008 |
| Naval | 0.001±0.0001 | **0.001±0.0000** | 0.001±0.0001 | 0.001±0.0001 | 0.006±0.0000 | 0.003±0.0001 | 0.010±0.0013 | 0.004±0.0011 |
| Power Plant | 4.182±0.0402 | 4.220±0.0744 | 4.112±0.0378 | 4.184±0.0591 | 4.124±0.0345 | **4.028±0.0347** | 4.065±0.0382 | 4.075±0.0366 |
| Protein | 4.539±0.0288 | 4.188±0.0313 | 4.014±0.0326 | **3.960±0.0110** | 4.688±0.0115 | 4.251±0.0153 | 4.094±0.0285 | 3.970±0.0376 |
| Wine | 0.645±0.0098 | 0.651±0.0108 | 0.652±0.0101 | 0.650±0.0158 | **0.635±0.0079** | 0.643±0.0077 | 0.641±0.0086 | 0.637±0.0079 |
| Yacht | 1.182±0.1645 | 1.542±0.1920 | 1.107±0.0863 | 1.265±0.1287 | 1.015±0.0542 | **0.848±0.0495** | 0.893±0.0991 | 1.711±0.2288 |
| Year | 8.932±NA | 8.976±NA | 8.933±NA | 9.045±NA | **8.869± NA** | 8.918±NA | 8.874±NA | 8.934±NA |

Finally, we have that

$$\frac{\partial \log Z}{\partial m_{i,j}^w} = \delta_j^m \frac{\partial m_i^a}{\partial m_{i,j}^w} + \delta_j^v \frac{\partial v_i^a}{\partial m_{i,j}^w}, \tag{21}$$

$$\frac{\partial \log Z}{\partial v_{i,j}^w} = \delta_j^m \frac{\partial m_i^a}{\partial v_{i,j}^w} + \delta_j^v \frac{\partial v_i^a}{\partial v_{i,j}^w}. \tag{22}$$

## 2 Results with neural networks including more than one hidden layer

We repeated the experiments from Section 5.1 in the main document for the methods BP and PBP, using neural networks with 2, 3 and 4 hidden layers. We used networks with 50 units in each hidden layer, except in the datasets *Year* and *Protein*, where we used 100. Table 1 shows the average test RMSE and the corresponding standard errors obtained by $PBP_x$ and $BP_x$, where $x$ is the number of hidden layers in the network. PBP has the best overall predictive performance, with $PBP_2$ achieving the best results in 5 datasest. Note that the optimal number of hidden layers in PBP is problem dependent. In datasets such as *Wine* and *Year* one single hidden layer is optimal, while in *Protein* we find that 4 hidden layers is better.

## 3 Error in the second approximation in equation (12) in the main text

In this section we evaluate the error in the second approximation performed in equation (12) in the main document. This approximation consists in replacing the Student's $t$ density with a Gaussian density that has the same mean and variance. This approximation becomes more and more accurate as the degrees of freedom in the Student's $t$ density increase. This will often be the case as we iterate over the data and we reduce our uncertainty on the value of the noise parameter $\gamma$. We evaluated the relative error in $\log Z$ caused by this approximation as PBP iterates over the data of the Boston Housing dataset in the experiments of Section 5.1 in the main document. The left plot in Figure 1 shows the error during the first 100 iterations of PBP over the individual datapoints. The right plot shows the error during the last 100 iterations of the method. We can see that the error is very small in the second case. In particular, at this stage we are highly confident on the value of the noise parameter $\gamma$ and the parameters $\alpha^\gamma$ and $\beta^\gamma$ in the posterior approximation take relatively high values. This increases the number of degrees of freedom of the Student's $t$ density in equation (12), what improves the quality of the Gaussian approximation.

## 4 List of approximations

In this section we list all the approximations performed by the method PBP. The list of approximations is

- We use expectation propagation (EP) to adjust a parametric approximation, given by equation (8) in the main document, to the exact posterior distribution, given by equation (3) in the main document.

- In our implementation of EP, we refine the parameters $\alpha^\gamma$, $\beta^\gamma$, $\alpha^\lambda$ and $\beta^\lambda$ of the posterior approximation by matching the first and second moments of $\lambda$ and $\gamma$. The KL divergence would be minimized by matching the expectation of the sufficient statistics of a Gamma distribution, but this does not have an analytical solution.

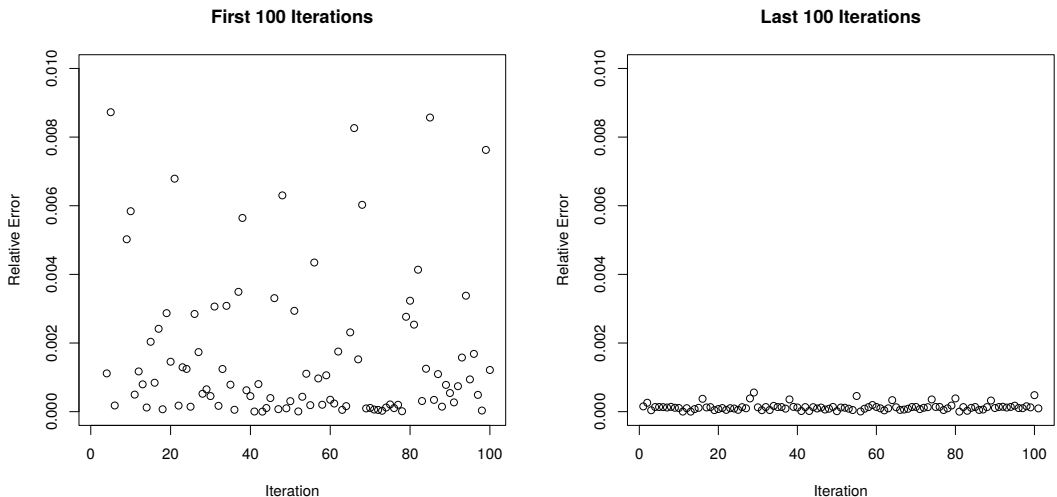

Figure 1: Relative error in the approximation of $\log Z$ as PBP iterates over the data of the Boston Housing dataset. in the experiments of Section 5.1 of the main paper. Left, relative error during the first 100 iterations of the method over the individual datapoints. Right, relative error during the last 100 iterations of the method. We can see that the error is very small during the last iterations. At this stage we are highly confident on the value of the noise parameter $\gamma$, the parameters $\alpha^\gamma$ and $\beta^\gamma$ in the posterior approximation take relatively high values and the degrees of freedom of the Student's $t$ density in equation (12) are high, what increases the quality of the Gaussian approximation.

- We approximate the normalization constants in equations (11) and (12) of the main document by replacing a Student's $t$ density with a Gaussian density that has the same mean and variance.

- EP requires to keep in memory one approximate factor for each exact factor in the numerator of the posterior distribution. With massive data the number of exact likelihood factors is very large and keeping in memory all the corresponding approximate factors is inpractical. To avoid this, we do not keep these approximate factors in memory and we do not remove them from the current approximation before processing each datapoint. This is equivalent to doing multiple ADF passes through the data, treating each likelihood factor as a novel example. A disadvantage of this approach is that it can lead to underestimation of the posterior variance when too many iterations are done over the data.

## 5 Derivations of equations (9) and (10) in the main text

Let the Gamma density be defined as $\text{Gamma}(x|\alpha,\beta) = \beta^\alpha x^{\alpha-1} \exp\{-x\beta\}\Gamma(\alpha)^{-1}$. We denote the normalization constant of $f(x)\text{Gamma}(x|\alpha,\beta)$ by $H(\alpha,\beta)$. In particular,

$$H(\alpha,\beta) = \int f(x)\text{Gamma}(x|\alpha,\beta)\,dx\,. \tag{23}$$

Note that we explicitly write $H$ as function of $\alpha$ and $\beta$. Then we have that the first and second moments of the normalized version of $f(x)\text{Gamma}(x|\alpha,\beta)$ are given by

$$\frac{1}{H(\alpha,\beta)}\int xf(x)\text{Gamma}(x|\alpha,\beta)\,dx = \frac{H(\alpha+1,\beta)\alpha}{H(\alpha,\beta)\beta}\,, \tag{24}$$

$$\frac{1}{H(\alpha,\beta)}\int x^2\text{Gamma}(x|\alpha,\beta)\,dx = \frac{H(\alpha+2,\beta)\alpha(\alpha+1)}{H(\alpha,\beta)\beta^2}\,. \tag{25}$$

4

Thus, each moment can be easily approximated given a procedure to approximate the normalization constant $H(\alpha, \beta)$. For this, we only have to substitute $H(\alpha, \beta)$, $H(\alpha + 1, \beta)$ and $H(\alpha + 2, \beta)$ in the previous expressions with their corresponding approximations. Note that the mean and variance of Gamma$(x|\alpha, \beta)$ are given by $\alpha/\beta$ and $\alpha/\beta^2$, respectively. We can then find the new parameters $\alpha^{\text{new}}$ and $\beta^{\text{new}}$ of a Gamma distribution that has the same mean and variance as the normalized version of $f(x)$Gamma$(x|\alpha, \beta)$ by solving the system of equations given by

$$\frac{\alpha^{\text{new}}}{\beta^{\text{new}}} = \frac{H(\alpha+1,\beta)\alpha}{H(\alpha,\beta)\beta}, \qquad \frac{\alpha^{\text{new}}}{[\beta^{\text{new}}]^2} = \frac{H(\alpha+2,\beta)\alpha(\alpha+1)}{H(\alpha,\beta)\beta^2} - \left\{\frac{H(\alpha+1,\beta)\alpha}{H(\alpha,\beta)\beta}\right\}^2, \qquad (26)$$

Let $Z = H(\alpha, \beta)$, $Z_1 = H(\alpha+1, \beta)$ and $Z_2 = H(\alpha+2, \beta)$. Then

$$\alpha^{\text{new}} = \left[ZZ_2 Z_1^{-2}(\alpha+1)/\alpha - 1.0\right]^{-1}, \qquad (27)$$

$$\beta^{\text{new}} = \left[Z_2 Z_1^{-1}(\alpha+1)/\beta - Z_1 Z^{-1}\alpha/\beta\right]^{-1}. \qquad (28)$$

## 6 EP updates for the approximate factors corresponding to the prior

The only prior factors that need to be processed multiple times using expectation propagation are the factors in equation (2) in the main document. The other Gamma priors on $\lambda$ and $\gamma$ have the same functional form as the posterior approximation $q$. This means that they need to be incorporated only once into $q$ since any removal and posterior re-incorporation of these factors would not produce any improvement in $q$.

We re-write here the expression for the prior factors than need to be processed multiple times, that is, equation (2) from the main document:

$$p(\mathcal{W}|\lambda) = \prod_{l=1}^{L} \prod_{i=1}^{V_l} \prod_{j=1}^{V_{l-1}+1} \mathcal{N}(w_{ij,l}|0, \lambda^{-1}). \qquad (29)$$

We also re-write here the expression for the posterior approximation $q$:

$$q(\mathcal{W}, \gamma, \lambda) = \left[\prod_{l=1}^{L} \prod_{i=1}^{V_l} \prod_{j=1}^{V_{l-1}+1} \mathcal{N}(w_{ij,l}|m_{ij,l}, v_{ij,l})\right] \text{Gamma}(\gamma|\alpha^\gamma, \beta^\gamma) \text{Gamma}(\lambda|\alpha^\lambda, \beta^\lambda). \qquad (30)$$

We denote each exact factor in (29) by

$$f_{ij,l}(w_{ij,l}, \lambda) = \mathcal{N}(w_{ij,l}|0, \lambda^{-1}). \qquad (31)$$

Each of these exact factors is approximated by a corresponding approximate factor given by

$$\tilde{f}_{ij,l}(w_{ij,l}, \lambda) = \mathcal{N}(w_{ij,l}|\tilde{m}_{ij,l}, \tilde{v}_{ij,l})\text{Gamma}(\lambda|\tilde{\alpha}_{ij,l}, \tilde{\beta}_{ij,l}), \qquad (32)$$

Initialliy all the $\tilde{f}_{ij,l}$ are uniform, that is, $\tilde{m}_{ij,l} = 0$, $\tilde{v}_{ij,l} = \infty$, $\tilde{\alpha}_{ij,l} = 1$ and $\tilde{\beta}_{ij,l} = 0$. EP starts to incorporate all the $f_{ij,l}$ into $q$ once it has already incorporated the Gamma priors for $\lambda$ and $\gamma$. The first time $f_{ij,l}$ is incorporated into $q$ we update $\tilde{f}_{ij,l}$ and $q$ as follows:

$$\tilde{m}_{ij,l} = 0, \qquad \tilde{v}_{ij,l} = \beta_0^\lambda/(\alpha_0^\lambda - 1), \qquad (33)$$

$$m_{ij,l} = 0, \qquad v_{ij,l} = \beta_0^\lambda/(\alpha_0^\lambda - 1), \qquad (34)$$

where $\alpha_0^\lambda$ and $\beta_0^\lambda$ are the parameters of the Gamma prior on $\lambda$. These rules guarantee the matching of means and variances on $w_{ij,l}$ after approximating the Student's $t$ density in equation (11) in the main document with a Gaussian that has the same mean and variance.

5

On successive iterations, we refine $\tilde{f}_{ij,l}$ by first removing this approximate factor from $q$ to obtain a cavity distribution. This cavity is computed as the ratio of $q$ and $\tilde{f}_{ij,l}$. The cavity marginal distribuion on $w_{ij,l}$ and $\lambda$ is therefore

$$q^{\backslash ij,l}(w_{ij,l},\lambda) = \mathcal{N}(w_{ij,l}|m^{\backslash ij,l}, v^{\backslash ij,l})\text{Gamma}(\lambda|\alpha_\lambda^{\backslash ij,l}, \beta_\lambda^{\backslash ij,l}), \tag{35}$$

where

$$v^{\backslash ij,l} = \left[v_{ij,l}^{-1} - \tilde{v}_{ij,l}^{-1}\right]^{-1}, \qquad m^{\backslash ij,l} = v^{\backslash ij,l}\left[m_{ij,l}v_{ij,l}^{-1} - \tilde{m}_{ij,l}\tilde{v}_{ij,l}^{-1}\right], \tag{36}$$

$$\alpha_\lambda^{\backslash ij,l} = \alpha^\lambda - \tilde{\alpha}_{ij,l} + 1, \qquad \beta_\lambda^{\backslash ij,l} = \beta^\lambda - \tilde{\beta}_{ij,l}, \tag{37}$$

After this, we update the parameters of $q$ to match moments between $q(w_{ij,l},\lambda)$ and the normalized version of $f(w_{ij,l},\lambda)q^{\backslash ij,l}(w_{ij,l},\lambda)$. For this, we use expression (11) in the main text to approximate the normalization constant of $f(w_{ij,l},\lambda)q^{\backslash ij,l}(w_{ij,l},\lambda)$. This last step is obtained by replacing $q$ in equation (11) in the main text with the cavity distribution. Equations (6), (7), (9) and (10) from the main document are then used to obtain the new parameters $m_{ij,l}$, $v_{ij,l}$, $\alpha^\lambda$ and $\beta^\lambda$ for the posterior aproximation. Finally, we update the parameters for the approximate factor $\tilde{f}_{ij,l}$ using

$$\tilde{v}_{ij,l} = \left[v_{ij,l}^{-1} - [v^{\backslash ij,l}]^{-1}\right]^{-1}, \qquad \tilde{m}_{ij,l} = \tilde{v}_{ij,l}\left[m_{ij,l}v_{ij,l}^{-1} - m^{\backslash ij,l}[v^{\backslash ij,l}]^{-1}\right], \tag{38}$$

$$\tilde{\alpha}_{ij,l} = \alpha^\lambda - \alpha_\lambda^{\backslash ij,l} + 1, \qquad \tilde{\beta}_{ij,l} = \beta^\lambda - \beta_\lambda^{\backslash ij,l}. \tag{39}$$



\end{document}